\documentclass[conference]{IEEEtran}
\usepackage[english]{babel}
\IEEEoverridecommandlockouts
\usepackage{cite}
\usepackage{amsmath,amssymb,amsfonts}
\usepackage{algorithmic}
\usepackage{graphicx}
\usepackage{textcomp}
\usepackage[dvipsnames]{xcolor}
\usepackage[export]{adjustbox}
\usepackage{booktabs}
\usepackage{url}
\usepackage{soul}
\usepackage{float}
\usepackage{hyperref}

\usepackage{microtype}

\if@mathematic
\def\vec#1{\ensuremath{\mathchoice
		{\mbox{\boldmath$\displaystyle\mathbf{#1}$}}
		{\mbox{\boldmath$\textstyle\mathbf{#1}$}}
		{\mbox{\boldmath$\scriptstyle\mathbf{#1}$}}
		{\mbox{\boldmath$\scriptscriptstyle\mathbf{#1}$}}}}
\else
\def\vec#1{\ensuremath{\mathchoice
		{\mbox{\boldmath$\displaystyle#1$}}
		{\mbox{\boldmath$\textstyle#1$}}
		{\mbox{\boldmath$\scriptstyle#1$}}
		{\mbox{\boldmath$\scriptscriptstyle#1$}}}}
\fi

\def\BibTeX{{\rm B\kern-.05em{\sc i\kern-.025em b}\kern-.08em
    T\kern-.1667em\lower.7ex\hbox{E}\kern-.125emX}}
\begin{document}

\title{
Predictively Combatting Toxicity in Health-related Online Discussions through Machine Learning
}
\author{\IEEEauthorblockN{Jorge Paz-Ruza, Amparo Alonso-Betanzos, Bertha Guijarro-Berdiñas and Carlos Eiras-Franco}

\IEEEauthorblockA{Universidade da Coruña, CITIC - LIDIA Group, Campus de Elviña s/n 15071 A Coruña\\
Email: \{j.ruza, amparo.alonso.betanzos, berta.guijarro, carlos.eiras.franco\}@udc.es}

\thanks{This research work has been funded by MICIU/AEI/10.13039/501100011033 and ESF+ (FPU21/05783), ERDF A way of making Europe (PID2019-109238GB-C22, PID2023-147404OB-I00), and by the Xunta de Galicia (ED431C 2022/44) with the European Union ERDF funds. CITIC, as Research Center accredited by Galician University System, is funded by ``Consellería de Cultura, Educación e Universidade from Xunta de Galicia", supported in an 80\% through ERDF Operational Programme Galicia 2021-2027, and the remaining 20\% by ``Secretaría Xeral de Universidades" (ED431G 2023/01)}
}

\maketitle

\begin{abstract}
    In health-related topics, user toxicity in online discussions frequently becomes a source of social conflict or promotion of dangerous, unscientific behaviour; common approaches for battling it include different forms of detection, flagging and/or removal of existing toxic comments, which is often counterproductive for platforms and users alike. In this work, we propose the alternative of combatting user toxicity predictively, anticipating where a user could interact toxically in health-related online discussions. Applying a Collaborative Filtering-based Machine Learning methodology, we predict the toxicity in COVID-related conversations between any user and subcommunity of \textit{Reddit}, surpassing 80\% predictive performance in relevant metrics, and allowing us to prevent the pairing of conflicting users and subcommunities.
\end{abstract}


\section{Introduction}

In recent years, the decentralisation of online communication and the expansion of social media has boosted the popularity of public health-related topics in online conversations; more than 80\% of US citizens use online media for medical consultations before visiting a doctor. Our reliance on digital media to inform itself about important issues such as health and disease implies an overriding need to ensure the good nature of conversations and information about these issues on the internet.

However, social media has also been characterised by a steep rise in misinformation and toxic behaviours, such as hate speech, bullying, etc. On public health issues, such as COVID, this toxicity can fuel social and political conflict, or promote unscientifically based dangerous behaviours \cite{CHIPIDZA2021102397}, \cite{mccauley2013h1n1}.

Among social networks, \textit{Reddit} is popular in work in medical AI, e.g. for early detection of self-injury and depression \cite{10.1145/3485447.3512129} or anorexia \cite{losada2019overview}. The hierarchical and decentralised structure made Reddit a hub of heated debate during the onset of the COVID pandemic, with over 200,000 related posts per day. On the other hand, Reddit is known for its high percentage of toxic interactions, made by more than 15 percent of users \cite{almerekhi2022investigating}, these being the most read and visited \cite{CHIPIDZA2021102397}. Surprisingly, more than 80\% of users vary their degree of toxicity between different Reddit subcommunities or \textit{subreddits}, indicating that the toxicity of a user in a subcommunity does not depend solely on the characteristics of the user or the \textit{subreddit}, but on the interaction of both \cite{almerekhi2022investigating}. 

Efforts to combat this toxicity during the disinformation episodes in the COVID pandemic focused on detecting toxic subcommunities or comments \cite{healthcare_coalition_2021} \cite{Detoxify} and banning users who exhibited these behaviours \cite{reddithate}. These reactive measures have become the most prevalent methods to combat toxicity in large-scale online platforms, and are supported either by professional or volunteer content moderators \cite{cao2024toxicity}. 

However, these detect-and-react moderation approaches come with a myriad of issues. Professional content moderation incurs high operational costs, recently causing platforms like \textit{X} (formerly \textit{Twitter}) or \textit{Facebook} to migrate to volunteer-centered moderation \cite{bbcMetaReplace}. Conversely, volunteer-based moderation is generally more susceptible to bias and under-moderation, depending on the platform's audience. Furthermore, both professional and volunteer-based reactive moderation of online toxicity has serious negative psychological effects on the moderating individuals \cite{spence2023psychological}, and incite toxic users to accumulate in radical communities with more potential for misinformation and public risk \cite{almerekhi2022investigating}. 

In the grand scheme, reactive approaches to combatting online toxicity only avoid critical situations, can be ineffective and biased, and involve the assistance of human operators; they do not solve the underlying problem, and instead tend to worsen it in the long term. As a shift in paradigm, we propose the alternative of combatting toxicity predictively: instead of detecting who and where \textit{has been} toxic, we propose to predict who and where \textit{will} be toxic. 

In this work, we explore the anticipation of toxic behaviour in COVID conversations in \textit{Reddit}, predicting the expected toxicity of any user in any subcommunity. Our approach models the latent toxicity characteristics of users and subcommunities using Matrix Factorization and Natural Language Processing and could allow, for example, to avoid recommending users to subcommunities where they are presumed to act in a toxic way, helping to reduce the occurrence of potentially dangerous online content on public health issues. The main contributions of this research are: 
\begin{itemize}
    \item The proposal of a novel approach to combatting online user toxicity; in contrast with classic reactive approaches, which are ineffective, resource-intensive and incite user radicalization, we introduce a \textit{predictive} approach to pre-emptively anticipate online user toxicity, which platforms could use to reduce its surfacing and the subsequent resources required to detect and moderate it.  
    \item The development of a Matrix Factorization-based Collaborative Filtering technique capable of predicting the future toxicity of a user when interacting with a new online subcommunity, in the fashion of the aforementioned toxicity approach. 
    \item The design of an adapted \textit{Leave Out Last Item} data partitioning method suitable for binary classification-oriented Collaborative Filtering tasks.
    \item The evaluation of our toxicity prevention algorithm in a real-world dataset of COVID-related discussions in \textit{Reddit}, achieving $>80$\% predictive performance in relevant metrics, proving our method can prevent the pairing of conflicting users and subcommunities. 
\end{itemize}

\section{Data Acquisition and Analysis}
\label{materials}

To work with real toxic and non-toxic user interactions, we collected comment data from \textit{Reddit} through the archiving platform \textit{PushShift} \cite{pushshift.io_2019}; suitable comment data for other platforms could not be obtained due to their recent decisions towards discontinuing API data access for research purposes \cite{10.1145/3589334.3645503}. The data acquisition, filtering and pre-processing process used, shown in Figure \ref{fig:datagathering}, is as follows:

\begin{figure}[h]
    \centering
    \includegraphics[width=\columnwidth]{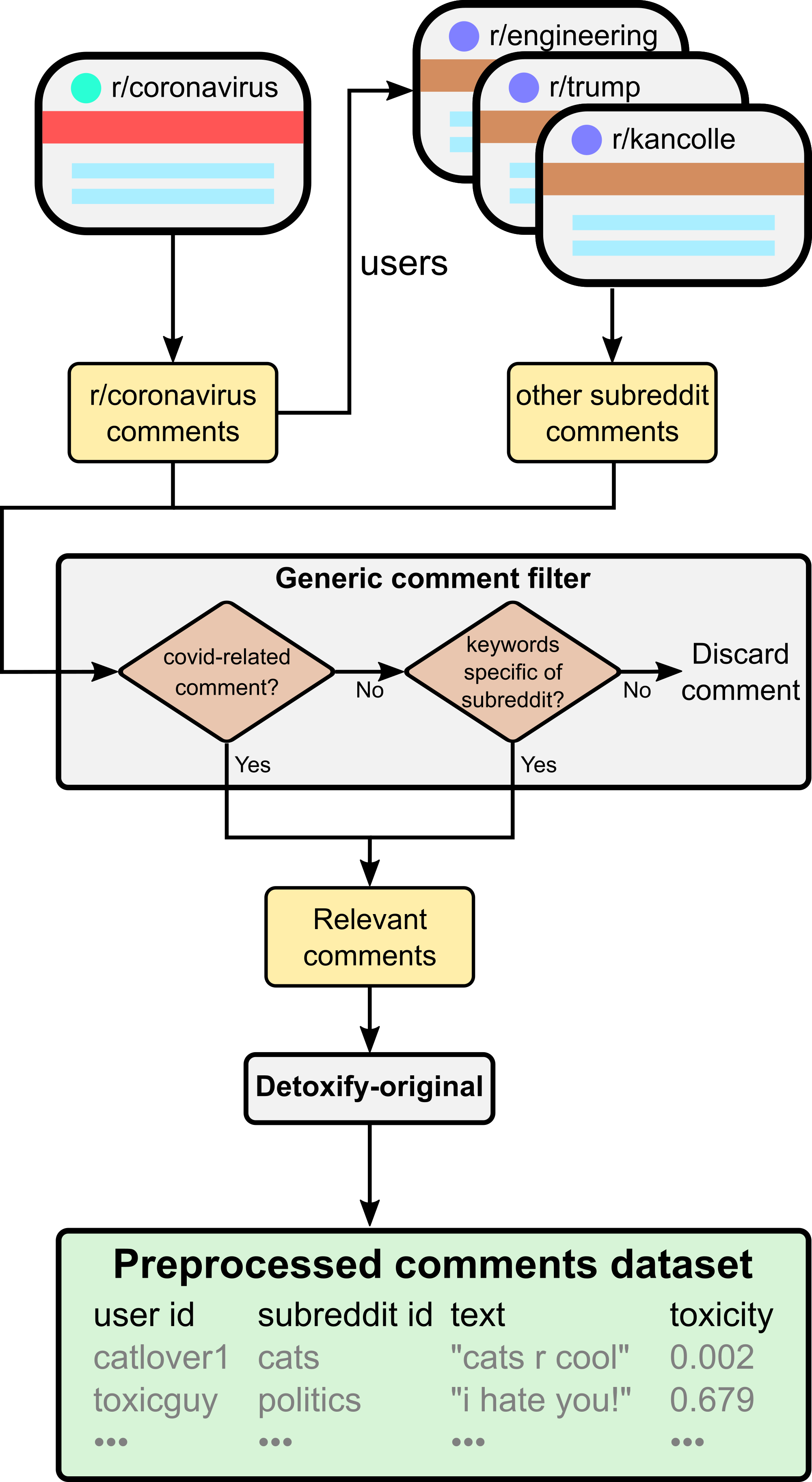}
    \caption{Acquisition, filtering and pre-processing processes for online \textit{Reddit} comment data.}
    \label{fig:datagathering}
\end{figure}

\begin{enumerate}
    \item We obtain comments posted on subreddit \textit{r/coronavirus} during March 2021 (approximating the start of worldwide COVID vaccination campaigns).

    \item For each user authoring comments obtained in 1), we extract up to 1000 of their comments across Reddit between January and June 2021 and add them to the set.

    \item We remove comments on subreddits where $\leq20$ users have commented, to discard those without sufficient conversation.

    \item We remove ``generic comments‘’ from the set, i.e. those unrepresentative of the subreddit they belong to and unrelated to COVID, using a keyword-based method for each subreddit $s$:

    \begin{enumerate}
        \item Identify $P(s)$, the set of the 100 most popular words in $s$. 
        \item Identify $P(\overline{s})$, the set of the 100 most popular words outside $s$.
        \item Calculate $S(s) = P(s) \setminus P(\overline{s})$, i.e. the popular words specific to $s$. 
        \item Tag comments as related to COVID if they contain keywords such as ‘COVID’, ‘vaccine’, ‘pfizer’, ‘lockdown’.... \footnote{Authors have temporarily removed this link to the work's repository to comply with double-blind review policies}     
        \item Label comments as ``generic‘’ if they do not contain any words from $S(s)$ and are not related to COVID. 
    \end{enumerate}
    
    \item Get the toxicity of each comment $tox(c) \in [0,1]$ using the \textit{Detoxify-original} language model  \cite{Detoxify}.
\end{enumerate}

This process yielded 625,673 comments posted by 25,893 users on 331 subreddits, which can be grouped into 175,531 unique user-subreddit interactions, on which we want to predict their toxicity. We observe several interesting features in user interactions on Reddit, based on the distribution of the data:

\begin{itemize}

    \item There is a large number of inactive or \textit{niche} users on Reddit, i.e. who post on few subreddits: each user interacts on average with $\approx6.77$ subreddits. This dispersion is typical in dyadic data (relating two entities) and negatively influences the training of Machine Learning models. 

    \item The majority of users do not post toxic comments when discussing health on Reddit, with 9.96\% of toxic comments in the aggregate, similar to previous work. Furthermore, as Figure \ref{fig:toxicityperinteraction} shows, a user's toxicity on a subreddit tends to be consistent (toxic or non-toxic, as indicated by the peaks in the distribution at toxicities 0 and 1), which suggests user toxicity in a given subreddit can be categorized in a binary manner.
    
    \begin{figure}[h]
    \centering
    \includegraphics[width=\columnwidth]{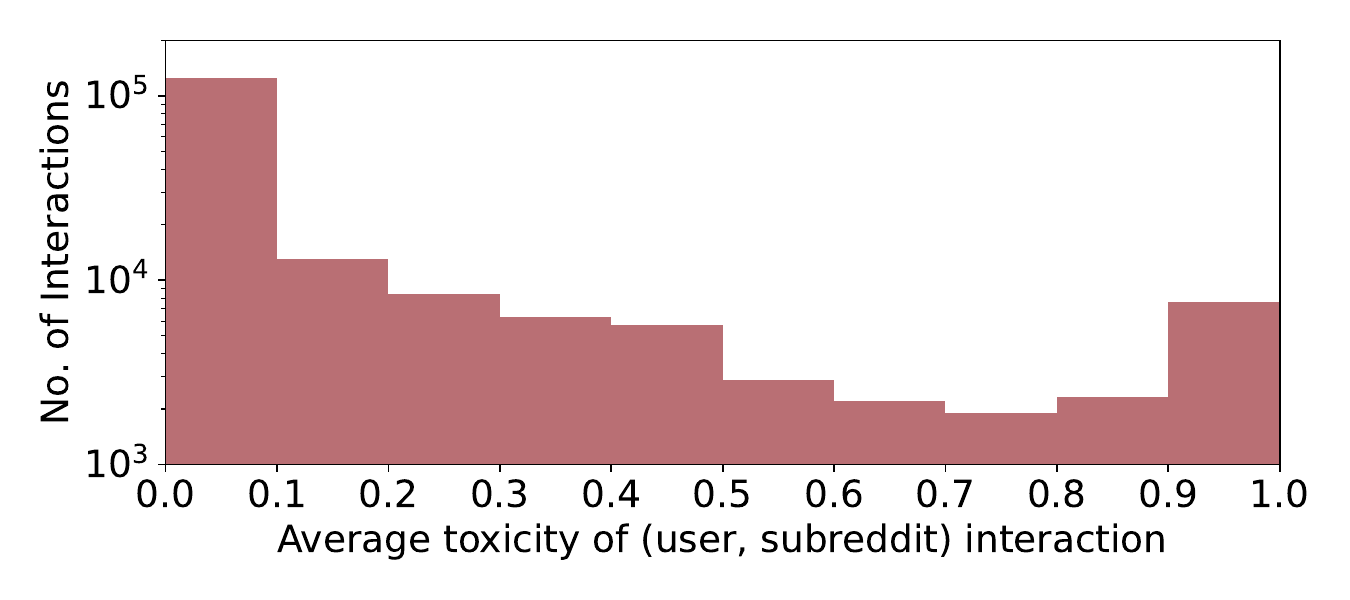}

    \caption{Distribution of average comment toxicity in each interaction (user, subreddit) in the set. Note the logarithmic scale on the y-axis.}
    \label{fig:toxicityperinteraction}
    \end{figure}    
\end{itemize}
\section{Methods}
\label{sec:toxicbert}

This section describes the Natural Language Processing and Matrix Factorisation techniques that allow identifying the toxicity of existing comments and using it to model the toxicity of future user-subreddit interactions, respectively.

\subsection{Detoxify}

To tag the toxicity of comments we use \textit{Detoxify-original} \cite{Detoxify}, a pre-trained language model. We do not re-train the model or fine-tune the model, and instead apply a pure Transfer Learning paradigm, taking the output of the pre-trained model for each comment $c$ (the probability of the label \textit{``toxic‘’}) as its toxicity $tox(c) \in [0,1]$.

Note that $tox(c)$ is the toxicity of an individual comment; to define the toxicity of an entire user-subreddit interaction, it is necessary perform an aggregation of the toxicities of the comments in that interaction.

\subsection{Matrix Factorization (MF)}
\label{sec:mf}

As anticipated in Section \ref{materials}, user-subcommunity interaction toxicity prediction involves the use of dyadic data, which relates entities of two types; in this case, this relation is the predicted toxicity of a given new user-subreddit interaction.

\begin{figure}[t]
    \centering
    \includegraphics[width=0.9\columnwidth]{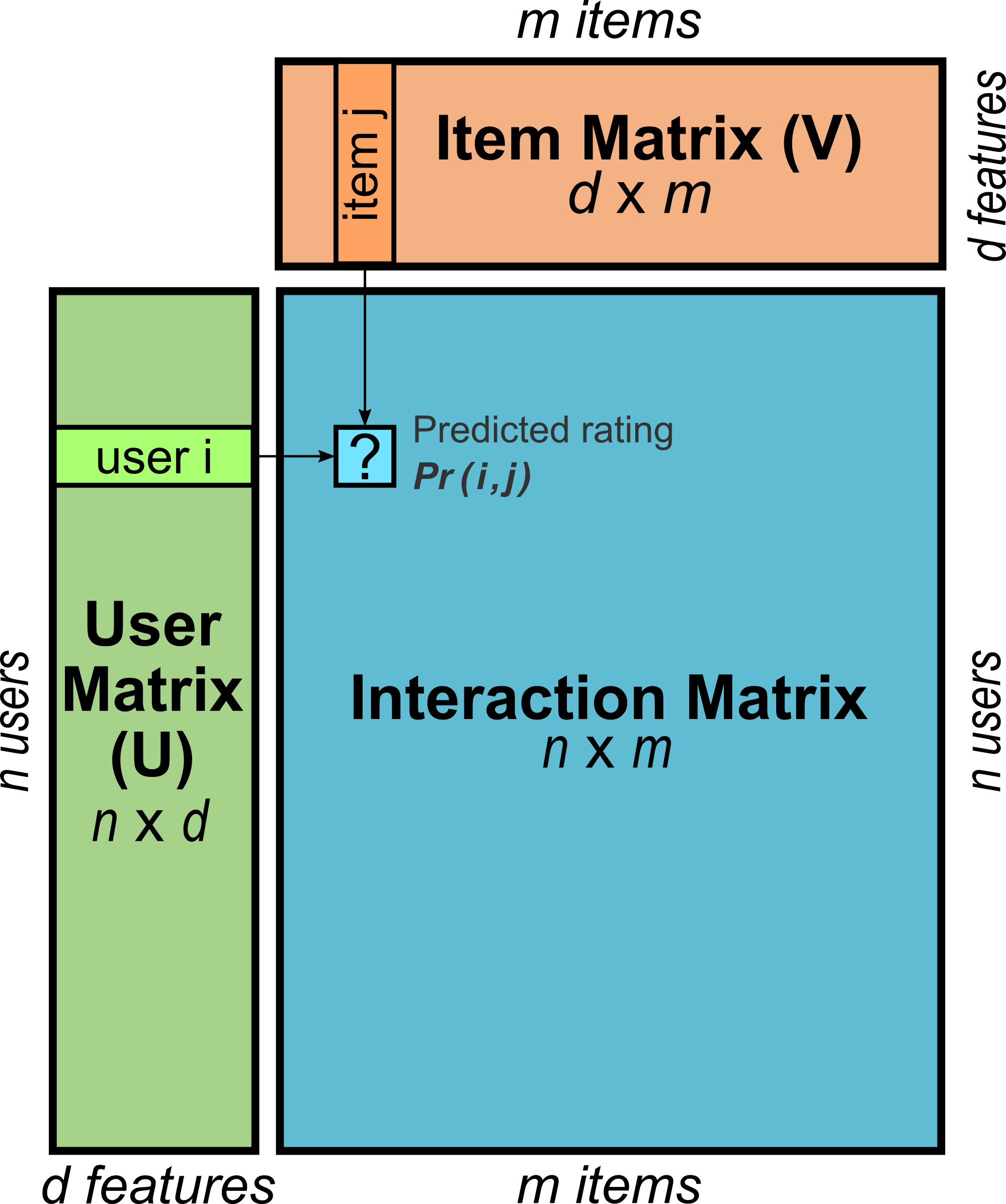}
    \caption{The Matrix Factorisation (MF) paradigm. The full interaction matrix is modelled as two low-dimensional matrices $U$ and $V$, representing the latent characteristics of users and items, respectively. The resulting value of any interaction can be obtained as the scalar product of a column of $V$ and a row of $U$.}
    \label{fig:mf}
\end{figure}

In a dyadic dataset consisting of interactions between $n$ users and $m$ items (corresponding to users and subreddits), the space of possible interactions can be represented as a matrix of size $n \times m$. This technique seeks to factorise this matrix into two matrices $U$ and $V$, of size $n \times d$ and $d \times m$ respectively, where $d \times m,n$ is the number of latent features. As shown in Figure \ref{fig:mf}, the result is two matrices of much smaller combined size than the original one ($(n+m)\times d$ vs. $n\times m$), where each column of $V$ represents the latent features of an item (here, the particularities of a subreddit) and each row of $U$ represents the preferences of a user according to those features (here, a user's toxicity patterns). 

This modelling of interactions using MF facilitates their use to predict the values of new interactions: using $U$ and $V$ these values (corresponding to the toxicity of the interaction, in our case) can be computed as the scalar product of two vectors (row and column) of size $d$.

Using MF to model a dyadic set requires choosing how to obtain the latent representations of users and items (columns of $V$ and rows of $U$). In the following, we detail our choice to obtain these representations following Collaborative Filtering (CF) approach, such that they are learned by a Machine Learning model implicitly from existing interactions.

\section{The proposed method}
\label{sec:proposal}

This section formally defines the classification task devised for user-subreddit toxicity prediction, the Machine Learning method proposed to solve it, and the evaluation policy designed to assess its performance. 

In the dyadic dataset of online conversations on Reddit, let $U$ and $S$ be the sets of users and subreddits present. This initial dataset presents data in the form of comments $c \in (u,s)$, posted by user $u \in U$ on the subreddit $s \in S$. As we introduced in Section \ref{sec:toxicbert}, we can calculate the toxicity $tox(c) \in [0,1]$ of each comment using the Detoxify language model \cite{Detoxify}.

Instead of only detecting and punishing the toxicity of existing interactions like common content moderation methods, which is ineffective and counterproductive in the long term, this work's proposal is to predict the toxicity of an unobserved interaction $C$ between a user $u$ and a subreddit $s$, defined by the $n$ comments posted by $u$ on $s$, such that $C(u,s) = \{c_1, c_2, ..., c_n\}$; consequently, we propose that toxic interactions can be anticipated and prevented from occurring in the first place. 

To make the results of this work easily interpretable and quantifiable for the scientific community, and taking advantage of the fact that toxicity within an interaction is consistently toxic or non-toxic (see Figure \ref{fig:toxicityperinteraction}), we pose this task as a binary classification: predict whether the interaction $C(u,s)$ will be toxic or non-toxic. 

At the labelling level, we compute this binary class as:

\begin{equation}
    tox(C(u,s))= round(\frac{\sum_{c \in C(u,s)}{tox(c)}}{|C(u,s)|})
\end{equation}

\noindent so $tox(C(u,s))$ will be 0 if the average toxicity of the comments of $u$ in $s$ does not exceed $0.5$, and vice versa.

\subsection{Dataset Partitioning}
\label{sec:datasetprep}

Predicting the future toxicity of interactions in our data requires learning the latent user and subreddit characteristics beforehand, which precludes the use of traditional partitioning techniques (e.g. classic holdout or cross-validation): they could create Test partitions with user interactions and subreddits that the model has not been trained on; this phenomenon is called \textit{cold start problem}. 

\begin{figure}[H]
    \centering
\includegraphics[width=.93\columnwidth]{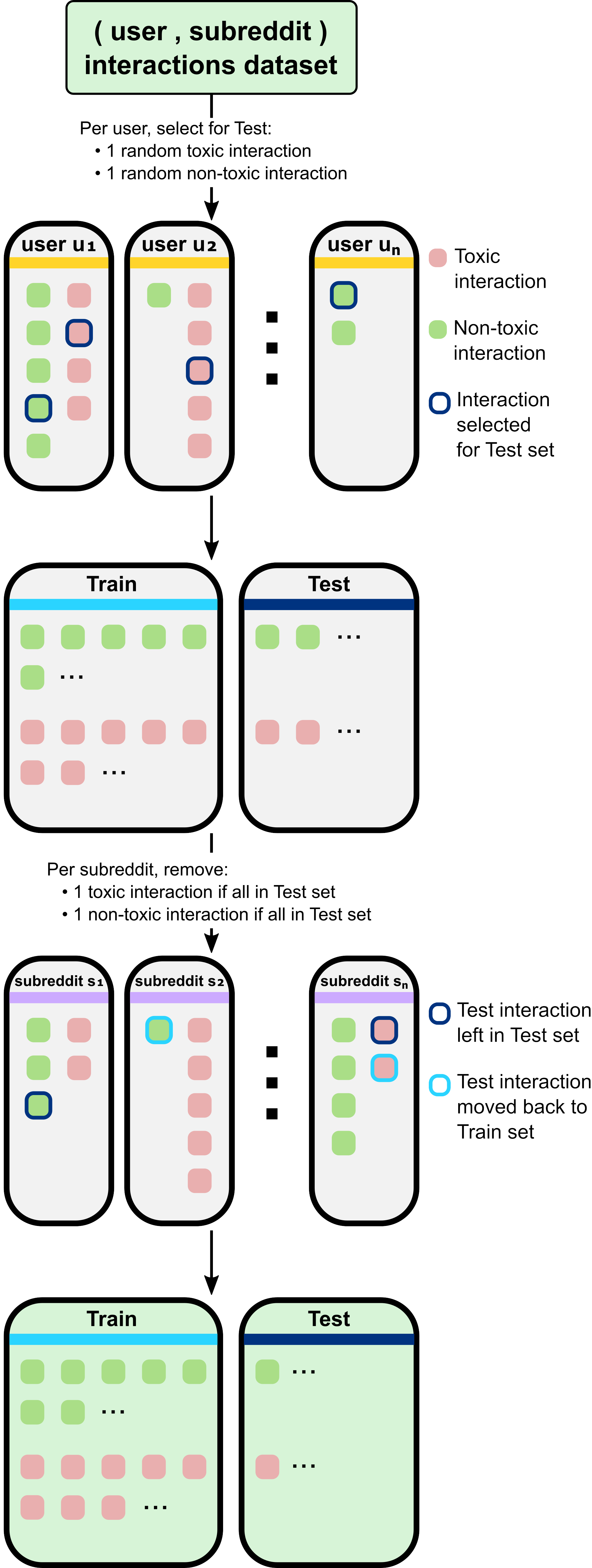}
    \caption{Adaptation of the LOLI partitioning strategy \cite{meng2020exploring} for binary classification on dyadic data. Note that this process is repeated to split the training set into the final training and validation sets.}
    \label{fig:datasplitting}
\end{figure}

In the State of the Art of dyadic data tasks \cite{meng2020exploring}, the most common partitioning technique is \textit{LOLI (Leave One Last Item)}, which reserves for the test set the time-wise last interaction (u, s) for each user with $\geq2$ interactions. However, this method is incompatible with our task, as our set does not have a unique temporal label per interaction, and LOLI is designed for tasks evaluated through rankings. Therefore, we propose a novel adaptation of the traditional LOLI method to binary ranking tasks (see Figure \ref{fig:datasplitting}) as follows:

 \begin{enumerate}

    \item For each user with $n\geq2$ toxic interactions, 1 of them is added to the test set and the remaining $n-1$ to the training set. 
    \item The process described in 1) is repeated, in this case with the non-toxic interactions.
    \item For each subreddit, if all its toxic interactions are in the test set, one of them is switched to the training set.
    \item The process described in 3) is repeated, in this case with the non-toxic interactions.
    \item Steps 1 to 4 are (optionally) repeated to split the training set into a final training set and a validation set.
\end{enumerate}

 Table \ref{tab:splitstats} shows, for the full set and the training, validation and test partitions, the total interactions, toxic and non-toxic, and the number of unique users and subreddits.

\begin{table}[htb]
\caption{Basic statistics of the Reddit COVID-related interaction data.}
\label{tab:splitstats}
\scriptsize
\resizebox{\columnwidth}{!}{ 
\begin{tabular}{lrrrrr}
\toprule
\textbf{} &
  \textbf{Interactions} &
  \textbf{\begin{tabular}[l]{@{}l@{}}Non-toxic\\ Interactions\end{tabular}} &
  \textbf{\begin{tabular}[l]{@{}l@{}}Toxic \\Interactions\end{tabular}} &
  \textbf{\begin{tabular}[l]{@{}l@{}}Unique\\ Users\end{tabular}} &
  \textbf{\begin{tabular}[l]{@{}l@{}}Unique\\ Subreddits\end{tabular}} \\ \midrule
\textbf{Full set}        & 175,531 & 158,578 & 16,953 & 25,893 & 331 \\
\textbf{Train}      & 123,076 & 111,850 & 11,226 & 25,893 & 331 \\
\textbf{Validation} & 23,954  & 22,220  & 1,734  & 22,314 & 331 \\
\textbf{Test}       & 28,501  & 24,508  & 3,993  & 24,577 & 331 \\ \bottomrule
\end{tabular}
}
\end{table}

\subsection{Model architecture}

\begin{figure}[htb]
    \centering
    \includegraphics[width=\columnwidth]{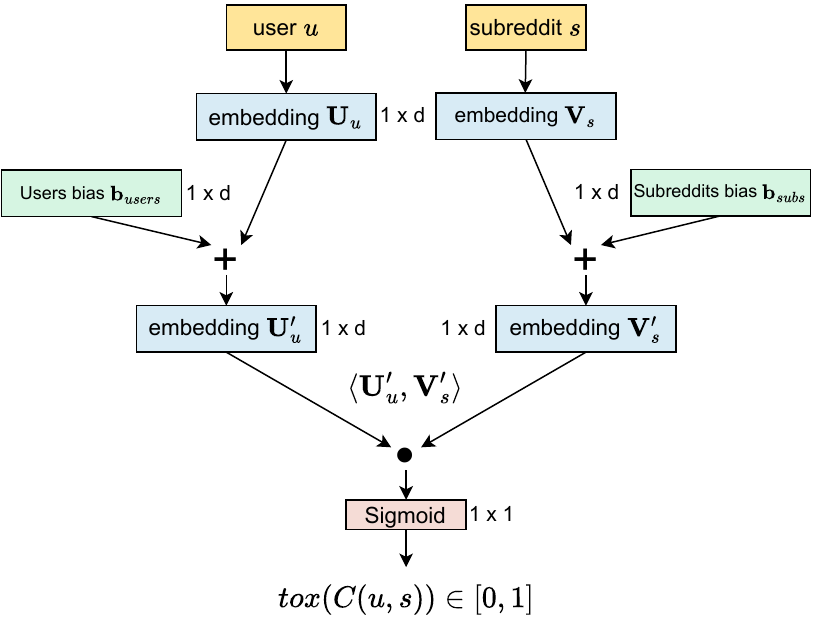}
    \caption{Topology of the Machine Learning model proposed to predict the toxicity of health-related conversations in unobserved user-subreddit interactions on the Reddit platform. The model receives the user identifiers $u$ and subreddit $s$, and predicts the expected toxicity $tox(C(u,s)) \in [0,1]$ from user behaviour on the subreddit. $d$ is the number of latent features used.}
    \label{fig:model_topology}
\end{figure}

To provide a modelling of our novel approach of predicting the toxicity of conversations in the future, unobserved interactions between users and subreddits, we have chosen a Matrix Factorisation (MF) architecture following the Collaborative Filtering paradigm; while more complex CF-based techniques exist, MF has been shown to achieve competitive state-of-the-art performance in CF tasks while constructing a simpler modelling of interactions \cite{10.1145/3383313.3412488}. Recalling that the input data includes the user and subreddit identifiers, and the output is the expected toxicity of the comments from the interaction between the two, we model the toxicity of the interaction as:

\begin{equation}
tox(C(u,s)) = \sigma( \langle \vec{U_u} + \vec{b_{users}} ,\vec{V_s} + \vec{b_{subs}} \rangle) 
\end{equation}

\noindent i.e. the scalar product of two latent vectors of user toxicity characteristics $\vec{U_u} \in \mathbb{R}^d$ and subreddit particularities $\vec{V_s} \in \mathbb{R}^d$. Prior to the scalar product, we apply two inductive bias vectors $\vec{b_{users}} \in \mathbb{R}^d$ and $\vec{b_{subs}}\in \mathbb{R}^d$ common to all users and subreddits, respectively; this modification does not add expressiveness to the model, but previous work and our experimentation have observed that it helps regularization and prevention of model over-fitting \cite{10.1145/3447548.3467376}. Finally, a sigmoidal activation is applied to restrict the values of $tox(C(u,s))$ to $[0, 1]$. Figure \ref{fig:model_topology} visually summarises the topology of this model.

To select the hyperparameters of the model we perform a grid search with the following search space: $d \in \{32, 64,128,256,512,1024,2048\}$, learning rate $\in {\{10^{-2}, 10^{-3}, 10^{-4}, 10^{-5}\}}$, L2 regularization \cite{ng2004feature} $\in \{{10^{-2}, 10^{-4}, 10^{-6}, 0}\}$, batch size $\in {\{256, 1024, 4096, 16384}\}$, and a maximum of 1000 epochs. We use an Adam optimizer \cite{kingma2014adam} with Binary Cross Entropy as loss function, as well as an early stopping policy \cite{prechelt_1998} with tolerance $\Delta=0.005$ and patience $p=7$ to prevent over-fitting. The best resulting model configuration used $d=128$, \textit{learning rate} $10^{-5}$, $L2=10^{-4}$ and batch size $1024$.

\subsection{Evaluation Methods}
\label{sec:evalmethods}

We assessed the predictive ability of our model and baselines using classical binary classification metrics: sensitivity, specificity, and geometric mean (G.Mean) of the class-wise sensitivities. Note that we avoided the use of plain accuracy as it is not suitable for imbalanced data; we also avoided the usage of AUC-ROC as it over-optimistic in heavily imbalanced datasets \cite{movahedi2023limitations}; lastly, we chose to use G.Mean instead of F1 Score, as the former is more suitable than the latter for imbalanced datasets when both classes have equal importance \cite{brownlee2020imbalanced}, such as in our proposed online toxicity prevention mechanism: both disallowing ``healthy'' users to interact and allowing toxic ones to do so will worsen the overall toxicity of the platform. 

As we are proposing a novel, predictive methodology for toxicity combat in online user-subcommunity interactions (in contrast with existing detect-and-react approaches), there are no existing state-of-the-art approaches that we may compare our designed CF-based model to. Consequently, we propose the following three baseline methods, in increasing order of complexity:

\begin{itemize}
    \item \textbf{NON}: All interactions are predicted as non-toxic, the majority class; this is equivalent to not performing a predictive anticipation of user-subcommunity toxicity, such as in current online platforms.
    \item \textbf{RND}: The interaction is predicted as toxic with probability $p$ equal to the proportion of toxic interactions in the training set ($p \approx 0.1)$, and vice versa.
    \item \textbf{USR}: For an interaction $C(u,s)$, the interaction is predicted as toxic with probability $p$ equal to the ratio of toxic interactions among the training interactions of $u$.  
\end{itemize}
\section{Results}
\label{sec:results}
This Section presents the results of our novel proposed method for combatting toxicity in online conversations, based on anticipating the future toxicity in unobserved user-subcommunity interactions; in particular, we test our method for the case of online COVID-related conversations in future user-subreddit interactions. 

Table \ref{tab:metrics} shows the performance of our method (MDL) and the three selected baseline methods (avg. and std. dev. of 5 runs). The best and second best result for each metric are highlighted in bold and underlined, respectively.

\begin{table}[h]
\caption{Test performance of our model (MDL) and baselines}
\label{tab:metrics}
\resizebox{\columnwidth}{!}{
\begin{tabular}{lcccc}
\hline
              & \textbf{Sensitivity} & \textbf{Specificity}                  & \textbf{G. Mean} \\ \hline
\textbf{NON}             & 0.000 $\pm$ 0.000                 & \textbf{1.000 $\pm$ 0.000} & 0.000 $\pm$ 0.000           \\
\textbf{RND}             & 0.100 $\pm$ 0.011                & 0.907 $\pm$ 0.014                                 & 0.301$ \pm$ 0.010           \\
\textbf{USR}              & \underline{0.241 $\pm$ 0.012}                & \underline{0.910 $\pm$ 0.004}                                 & \underline{0.468 $\pm$ 0.010}           \\
\textbf{MDL}              & \textbf{0.849 $\pm$ 0.023}         & 0.810 $\pm$ 0.017                                 & \textbf{0.829 $\pm$ 0.018}                \\ \hline
\end{tabular}
}
\end{table}

Our model obtained the highest geometric mean ($\sim83\%$) of class-wise sensitivities, a high performance for the difficulty of the task and considerably better than the three baselines. Analysing the sensitivity per class individually, we observe that, although its specificity is lower than that of the baselines, this is because the baselines are biased to predict that interactions are non-toxic, which is closer to existing detect-and-react approaches against toxicity: our method is the only one that is correctly able to discern toxicity in future user-subreddit interactions, with a sensitivity 4 times higher than that of the baselines. 

Based on the results, we can conclude that the proposed Machine Learning methodology is able to reliably predict toxicity in the outcome of user-subreddit interactions, and our predictive approach can be useful to prevent future toxic interactions. Our approach performs better than non-machine learning based methods and avoids class-wise biases or over-fitting of the model. 

\section{Conclusions and Future Work}

In this article, we introduced a Machine Learning model that predicts the toxicity of future interactions between any user and online subcommunity discussing public health issues. 

With this research, we contribute to the state of the art of combating toxicity and misinformation in online health conversations; in contrast to classical, purely detection-based approaches, which are ineffective and counterproductive in the long term, our methodology focuses on preventing and predicting toxic behaviours. For this, we use Matrix Factorisation-based Collaborative Filtering techniques that allow us to learn the latent ``toxicity characteristics" of each user and subreddit and predict their behaviour in future conversations. 

In this new proposed task of binary classification of toxicity in new user-subreddit interactions, our proposal achieves high sensitivity ($\sim80\%$) in both classes (toxic and non-toxic) on a large set containing user-subreddit interactions about COVID on Reddit. In order to evaluate this task, we also propose a new adaptation of the LOLI partitioning strategy that allows its use in binary classification problems on dyadic data without temporal information.

We identify different avenues of future work. First, it would be beneficial to test the presented methodology on other online media that contain conversations about public health, such as X (formerly Twitter), online articles, YouTube comments, etc. as well as other diseases, fitness, or mental health; however, as previously outlined in this work, recent movements towards data access restrictions by platforms have complicated efforts towards this goal. Alternatively, one option to seek greater performance of our methodology in the future is to incorporate textual or temporal information into the model; however, achieving this is challenging as each user-subcommunity interaction can consist of several comments with  different timestamps, making it difficult to synthesise meaningful textual or temporal information, but it could increase the approach's ability to infer latent user and subcommunity characteristics to prevent future user toxicity.

\bibliographystyle{ieeetr}
\bibliography{TOXICITY}{}

\begin{thebibliography}{10}
\bibitem{reddithate}
``R/redditsecurity - understanding hate on reddit, and the impact of our new policy.''
\end{thebibliography}

\end{document}